\documentclass[letterpaper, 10 pt, conference]{ieeeconf}  

\IEEEoverridecommandlockouts                              

\usepackage{graphics} 
\usepackage{graphicx}
\usepackage{amsmath} 
\usepackage{amssymb}  
\usepackage{acro}
\usepackage[style=ieee,sorting=none]{biblatex}
\usepackage{cleveref}
\creflabelformat{equation}{#2\textup{#1}#3}
\Crefname{figure}{Fig.}{Figs.}
\Crefname{equation}{Eq.}{Eqs.}
\Crefname{table}{Tab.}{Tabs.}
\usepackage{multirow}
\usepackage{stfloats}
\input{acronyms.acn}

\title{\LARGE \bf
An End-to-end Flight Control Network for High-speed UAV Obstacle Avoidance based on Event-Depth Fusion
}

\author{Dikai Shang, Jingyue Zhao$^*$, Shi Xu, Nanyang
Ye, Lei Wang$^*$
\thanks{$^*$Corresponding authors}
\thanks{Dikai Shang and Nanyang Ye are with Shanghai jiaotong University, Shanghai 200240, China
{\tt\small dkshang2025@sjtu.edu.cn, ynylincoln@sjtu.edu.cn}}
\thanks{Jingyue Zhao, Shi Xu and Lei Wang are with Defense Innovation Institute, AMS, Beijing 10071, China
{\tt\small jingyue-zhao@foxmail.com, xushi9018@aliyun.com,  leiwang@nudt.edu.cn}}
}

\begin{document}

\maketitle
\thispagestyle{empty}
\pagestyle{empty}

\begin{abstract}

Achieving safe, high-speed autonomous flight in complex environments with static, dynamic, or mixed obstacles remains challenging, as a single perception modality is incomplete. Depth cameras are effective for static objects but suffer from motion blur at high speeds. Conversely, event cameras excel at capturing rapid motion but struggle to perceive static scenes. To exploit the complementary strengths of both sensors, we propose an end-to-end flight control network that achieves feature-level fusion of depth images and event data through a bidirectional crossattention module. The end-to-end network is trained via imitation learning, which relies on high-quality supervision. Building on this insight, we design an efficient expert planner using Spherical Principal Search (SPS).
This planner reduces computational complexity from $O(n^2)$ to $O(n)$ while generating smoother trajectories, achieving over 80\% success rate at 17 m/s—nearly 20\% higher than traditional planners. Simulation experiments show that our method attains a 70–80\% success rate at 17 m/s across varied scenes, surpassing single-modality and unidirectional fusion models by 10–20\%. These results demonstrate that bidirectional fusion effectively integrates event and depth information, enabling more reliable obstacle avoidance in complex environments with both static and dynamic objects.

\end{abstract}

\section{INTRODUCTION}

\ac{UAVs} are increasingly used for search and rescue, logistics, and infrastructure inspection, thanks to their flexibility and agility~\cite{commercial2015,review2024tang,application2022li,multiuav2023ivic}. However, achieving safe and reliable autonomous flight in complex real‑world environments with static, dynamic, and mixed obstacles remains a critical research challenge~\cite{computer2021kakaletsis,autonomous2016bagloee,survey2000bishop}.

Multimodal fusion is widely considered promising for obstacle avoidance~\cite{kinematics2022bearzot,autonomous2023liang}, but most existing systems still rely on a single sensor modality or process information from different modalities independently. Depth cameras, which provide dense geometric reconstruction and precise distance estimation, are effective at detecting static obstacles. Consequently, depth-based methods for static obstacle avoidance have been extensively studied and applied. Early work by Antonio et al. introduced an end-to-end imitation learning framework for \ac{UAVs} obstacle avoidance using depth images, achieving flight speeds up to 7 m/s in cluttered, static environments like forests and buildings~\cite{learning2021loquercio}. Florence et al. later proposed a mapless method relying solely on instantaneous depth measurements, which reduced sensitivity to state estimation errors and enabled obstacle avoidance at speeds up to 10 m/s in forests~\cite{integratedflorence}. Lu et al. presented a single-stage, learning-based planner that integrates perception and motion planning within a unified network~\cite{you2024lu}. By combining reinforcement and imitation learning, this planner achieved reliable avoidance at 7 m/s in similar environments. However, the limited frame rate of depth cameras can cause motion blur during high-speed flight, and their narrow dynamic range reduces performance under challenging lighting. As a result, depth-based methods are typically confined to low-speed operation—often below 10 m/s—and struggle to perform reliably in dynamic or high-speed scenes.

Event cameras, or Dynamic Vision Sensors (DVS)~\cite{128times2008lichtsteiner,2003lichtsteiner}, are attracting growing interest for high-speed UAV navigation due to their microsecond-level latency and robustness to motion blur~\cite{eventbased2022gallego}. Unlike standard cameras, they do not capture full frames; instead, they generate asynchronous streams of events in response to pixel-level brightness changes~\cite{experimental2021holesovsky,128times2008lichtsteiner,deblur2025low}. This makes them exceptionally efficient at detecting moving objects. However, during obstacle avoidance, the UAV’s motion generates numerous events from the static background, which significantly complicates obstacle detection. To address this, existing approaches generally fall into two main categories. One filters out ego‑motion events to isolate independently moving objects—effective for dynamic obstacles, but unable to perceive static ones, making it unsuitable for static obstacle avoidance. The other recovers depth from events to support static avoidance; however, its dependence on reconstructing depth images makes it prone to motion blur in high‑speed scenarios, severely limiting its ability to handle dynamic obstacles at high speed.

In the first category, Valles et al. developed a fully neuromorphic perception-to-control pipeline for autonomous drone flight, employing spiking neural networks to translate event data directly into low-level control commands~\cite{fully2024paredes}. Similarly, Zhang et al. used a event-based detector with spiking neural networks to identify a single moving obstacle (a basketball) and trigger avoidance actions through a lightweight decision module~\cite{dynamic2023zhang}. Such approaches typically require distinguishing events caused by independently moving objects from those generated by the UAV’s own motion against a static background. However, avoiding static obstacles with DVS remains a challenge. 

In the second category, Anish et al. proposed the first method for quadrotors to perform static obstacle avoidance using only a monocular event camera. Their core idea is to predict depth as an intermediate output within an end-to-end network that maps directly to control commands~\cite{monocular2024bhattacharyaa}. Although event cameras offer significant advantages for high-speed UAV navigation, existing approaches are limited to either dynamic or static obstacle avoidance, and none effectively address mixed scenes where both types of obstacles appear.

In summary, vision-based obstacle avoidance research can be broadly categorized into two approaches. Systems based on depth cameras, which provide dense geometric information, are particularly effective for detecting and avoiding static obstacles. On the other hand, systems based on event cameras leverage their capacity to capture rapid intensity changes and are therefore typically employed for fast-moving dynamic obstacles. These complementary characteristics suggest that fusing the two modalities is a promising direction~\cite{monocular2024bhattacharyaa}.

While recent works have explored combining depth and event sensors for UAV obstacle avoidance, they typically employ decision-level fusion—processing each modality independently and merging results at the task level\cite{lowlatency2025chen,fastdynamicvision2021hea}. In contrast, we argue that feature-level fusion is essential for high-speed flight, as it allows the network to learn joint representations that capture the complementary nature of structural depth and motion-sensitive event data. Therefore, we propose an end-to-end UAV control network that achieves feature-level fusion of depth images and event data via a bidirectional crossattention module. The network is trained using imitation learning, a method whose performance critically depends on the quality of expert demonstrations. Therefore, we first design an efficient \ac{SPS} planner to enhance demonstration quality. This planner reduces the computational cost of trajectory planning from $O(n^2)$ to $O(n)$, enabling real-time generation of smoother, more stable reference trajectories and, consequently, a higher success rate in obstacle avoidance.

Our contributions are as follows:
\begin{itemize}
    \item  We introduce an end-to-end imitation learning network that achieves feature-level fusion of depth images and event data via a bidirectional crossattention module, directly mapping multimodal perceptions to control commands.
    \item We design an efficient \ac{SPS} expert planner, which reduces the computational complexity of trajectory search from $O(n^2)$ to $O(n)$. This planner generates smoother trajectories and achieves an obstacle avoidance success rate of over 80\% at 17 m/s across various scenarios, outperforming traditional expert methods by nearly 20\%.
    \item Extensive simulations show that our method consistently outperforms single-modality networks and unidirectional fusion baselines, achieving success rates of 70–80\% (a 10–20\% improvement) at 17 m/s in cluttered environments and demonstrating superior robustness.
\end{itemize}

\section{Method}

\subsection{Overall Imitation Learning Framework}

\begin{figure}[htbp]
  \centering
  \includegraphics[width=0.9\linewidth]{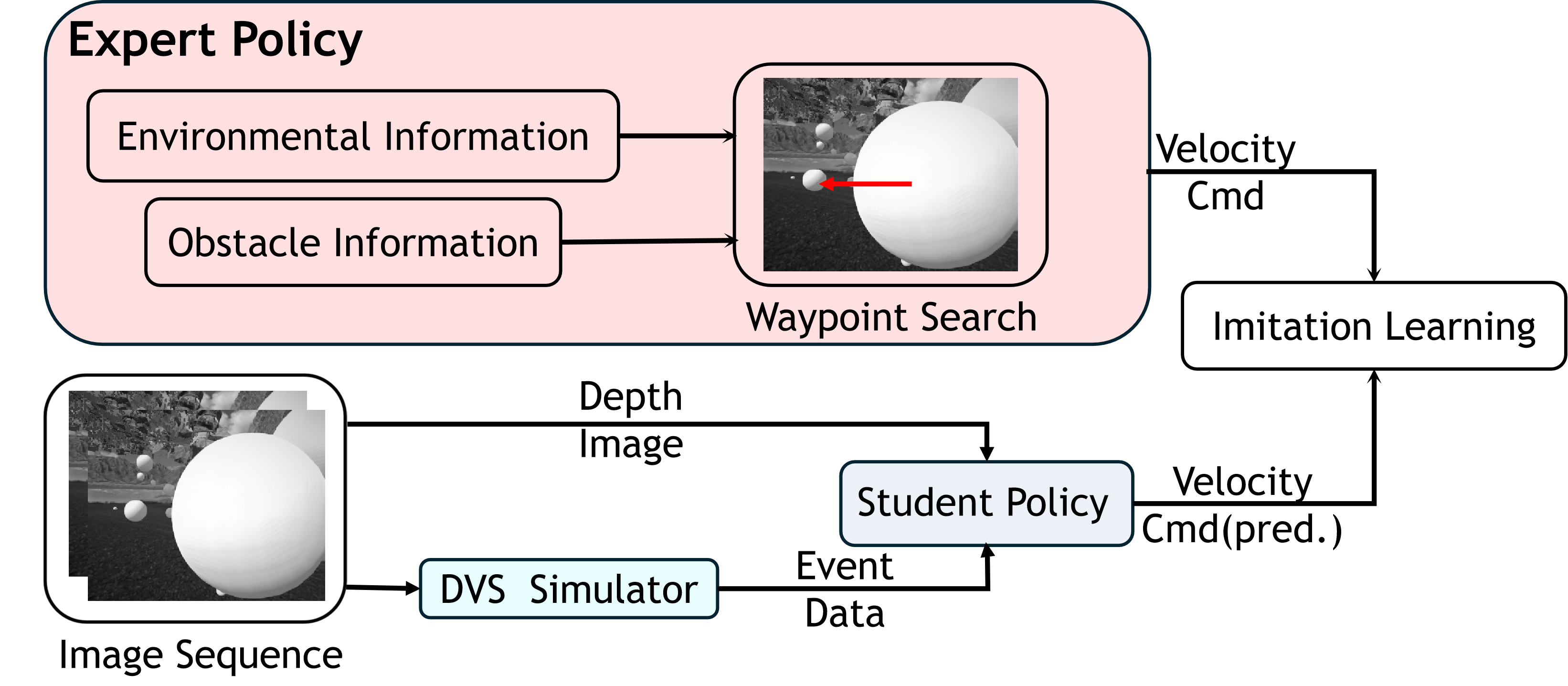}
  \caption{Overall model architecture. The expert planner generates optimal trajectories using global environmental information, while the student network learns end-to-end control from local sensor inputs (depth images and event data) and the UAV’s state.}
  \label{fig:overall_model_architecture}
\end{figure}

We present a robust end-to-end imitation learning framework designed for high-speed visual obstacle avoidance. As illustrated in \Cref{fig:overall_model_architecture}, the proposed architecture operates under a student-expert paradigm, comprising two distinct components: an Efficient Expert Policy and an End-to-end Student Policy.

\textbf{Efficient Expert Policy}: In the training phase, the expert serves as the supervisor. Using privileged information (environmental and obstacle data), it applies our proposed \ac{SPS} algorithm to produce optimal, smooth, and collision-free trajectories. These demonstrations supply the student network with the ground-truth supervision signals required for learning.

\textbf{End-to-end Student Policy}: The student policy learns to imitate the expert via an end-to-end network. To simplify the learning problem and ensure operational safety, the network’s output is defined as a normalized velocity command constrained to the xy-plane: $\tilde{\textbf{v}}_{pred} = (v_x, v_y, 0)$, where $v_x^2 + v_y^2 = 1$. This formulation explicitly prevents potentially hazardous vertical accelerations. Finally, the normalized command $\tilde{\textbf{v}}$ is scaled by a predefined desired flight speed $v_{set}$ to produce the final target velocity: $\textbf{v} = v_{set} \tilde{\textbf{v}}$.

\subsection{Efficient Expert Policy} \label{subsec:expert_policy}

\begin{figure}[htbp]
      \centering
  \includegraphics[width=0.9\linewidth]{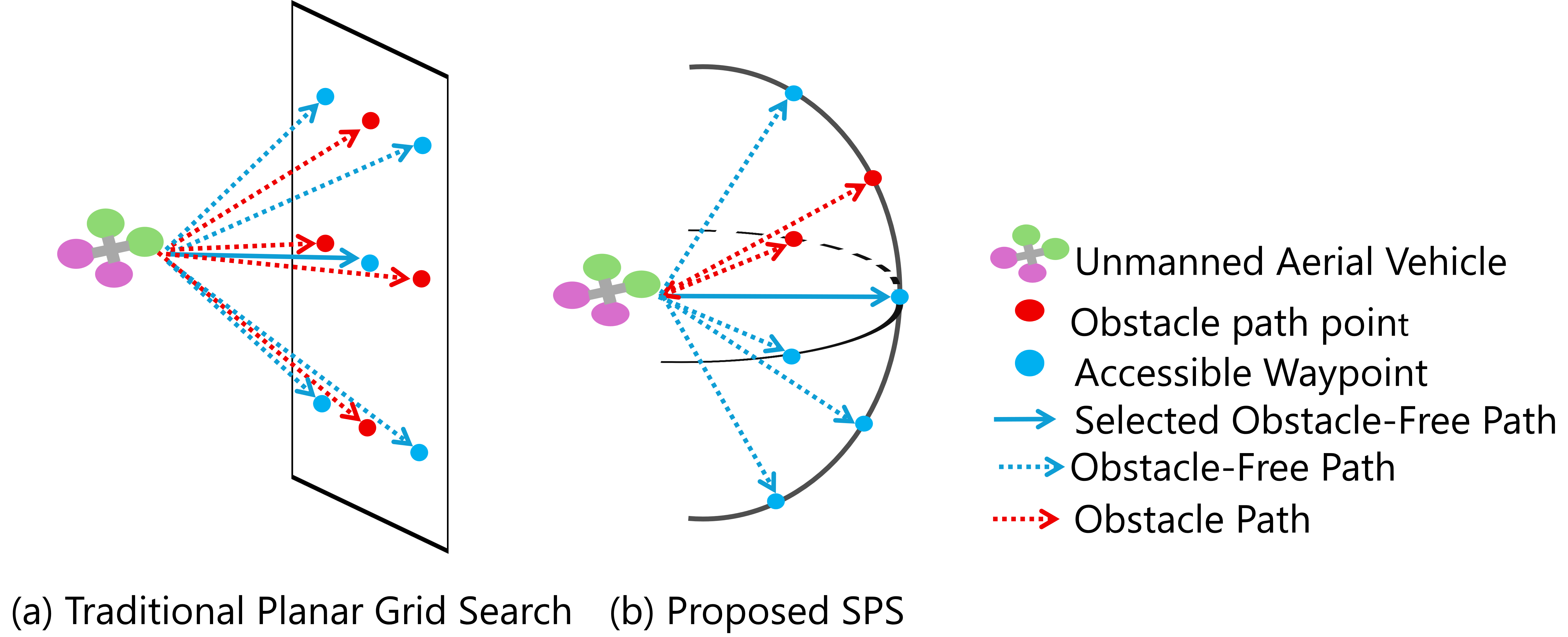}
  \caption{Comparison of path search strategies. (a) Traditional planar grid search; (b) Proposed \ac{SPS}.}
  \label{fig:expert_search_strategies}
\end{figure}

Traditional planners typically sample waypoints uniformly across a 2D plane shown on the (\Cref{fig:expert_search_strategies}(a)). With $n$ samples along each of the two orthogonal axes, the total number of waypoints is $n^2$, leading to $O(n^2)$ computational complexity. Moreover, since the sampled points are not required to follow the UAV’s principal directions of motion (up, down, left, and right), the resulting trajectories often involve unnecessary heading changes and are generally inefficient.

\begin{figure*}[htbp]
  \centering
  \includegraphics[width=0.8\linewidth]{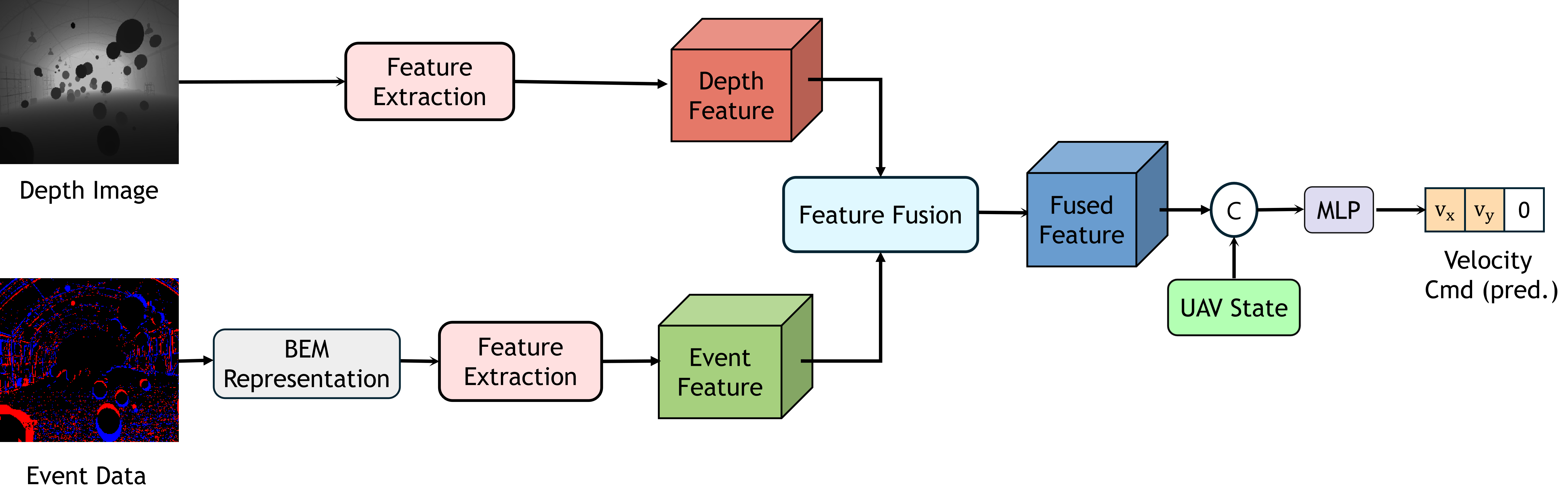}
  \caption{Overall student network architecture} 
  \label{fig:overall_student_network_architecture}
\end{figure*} 

We observe that, under constant-speed flight, all reachable positions within a given time window lie on a spherical surface, and any flight direction can be expressed as a vector sum of the UAV’s four principal axes. These two insights point to a more efficient sampling strategy: instead of searching exhaustively over a 2D grid, we sample waypoints only along the four principal directions on the sphere of reachable positions shown on the (\Cref{fig:expert_search_strategies}(b)). Inspired by these insights, we propose \ac{SPS}, which achieves linear time complexity $O(n)$ by sampling $\frac{n}{2}$ points along each of the four principal directions on the sphere. Moreover, since these directions align with the UAV’s natural axes of motion, the resulting trajectories are inherently smoother and more efficient—directly addressing the core limitation of conventional 2D planners. The detailed procedure of \ac{SPS} is as follows.

\textbf{Spherical Sampling \& Path Generation}: In accordance with the UAV’s designated speed $v_{set}$, candidate waypoints are sampled on a spherical surface of radius $r=v_{set}$ positioned ahead of the vehicle. Each waypoint is expressed in spherical coordinates $(r,\phi,\theta)$, where $\phi \in [-\pi/15,\pi/15]$ defines the allowable steering deviation and $\theta$ denotes the azimuth angle. The sampling scheme adapts to environmental geometry: for spherical obstacles, $\theta$ is selected from the discrete set $\{-\pi/2,0,\pi/2,\pi\}$; in forested scenes, $\theta$ is constrained to $\{0,\pi\}$. Finally, all spherical coordinates are converted to Cartesian space via the transformation given in \Cref{eq:coord_transform}.

\begin{equation}
\begin{aligned} \label{eq:coord_transform}
    x&=r \cos \phi \\
    y&=r \sin \phi \cos \theta \\
    z&=r \sin \phi \sin \theta
\end{aligned}
\end{equation}

\textbf{Feasibility Assessment}: Each candidate path undergoes collision detection against spherical obstacle models as illustrated in \Cref{fig:expert_search_strategies}. A path is deemed feasible (marked in blue) if its minimum distance to any obstacle exceeds the safety margin $d_{safety}$; otherwise, it is infeasible (marked in red).

\textbf{Optimal Path Selection}: From among the feasible paths, the optimal trajectory is selected by maximizing a weighted multi-criteria scoring function (\Cref{eq:score_function}). This function synthesizes several key objectives: goal alignment $s_{goal}$, calculated via normalized cosine similarity with the target direction; safety clearance $s_{safety}$, derived from the normalized distance to the nearest obstacle; lateral deviation $s_{x, offset}$, modeled as $1 - x_{offset}$; and longitudinal penalty $s_{y, offset}$, which applies a factor of 1.2 during turns to promote flight stability. The final score is obtained by combining these terms with the respective weights $w_{goal}=0.2$, $w_{safety}=0.4$, and $w_{y, offset}=0.4$.

\begin{equation}
\begin{aligned} \label{eq:score_function}
score = &w_{goal}\cdot s_{goal} + w_{safety} \cdot s_{safety}  \\
        &+ w_{x,offset} \cdot s_{x} + s_{y,offset}
\end{aligned}
\end{equation}

If no path meets the safety distance constraint $d_{safety}$, this threshold is progressively relaxed in increments of 0.2 until a feasible command is generated.

\subsection{End-to-end Student Network}
The end-to-end architecture of the student network, illustrated in \Cref{fig:overall_student_network_architecture}, is designed to efficiently fuse depth image and event data for robust control under resource constraints. Its pipeline comprises the following key components:

\textbf{BEM Representation}: In the absence of a physical event camera in simulation, event streams are synthesized from grayscale images using the Vid2e simulator~\cite{video2020gehrig}. These asynchronous streams are then divided into fixed-time windows, each of which is encoded as a Binary Event Mask (BEM, \Cref{eq:BEM_represent}, as shown in \Cref{fig:overall_student_network_architecture} BEM Representation), capturing local polarity imbalances within each window as a binary image.

\begin{align} \label{eq:BEM_represent}
    \mathbf{B E M}_{x, y}=\mathbf{1}\left(\left(\sum_{e \in E} \mathbf{1}\left(e_p=+1\right)-\sum_{e \in E} \mathbf{1}\left(e_p=-1\right)\right) \neq 0\right)
\end{align}

\textbf{Lightweight Visual Feature Extraction}: Considering the limited onboard computational resources of \ac{UAVs}, we employ MobileNet-V3-Small with the classification head removed—a lightweight network pretrained on ImageNet—as our feature extraction backbone (the Feature Extraction part in the \Cref{fig:overall_student_network_architecture}).
\begin{figure*}[htbp]
  \centering
  \parbox[t]{0.3\linewidth}{
    \centering
    \includegraphics[width=0.8\linewidth]{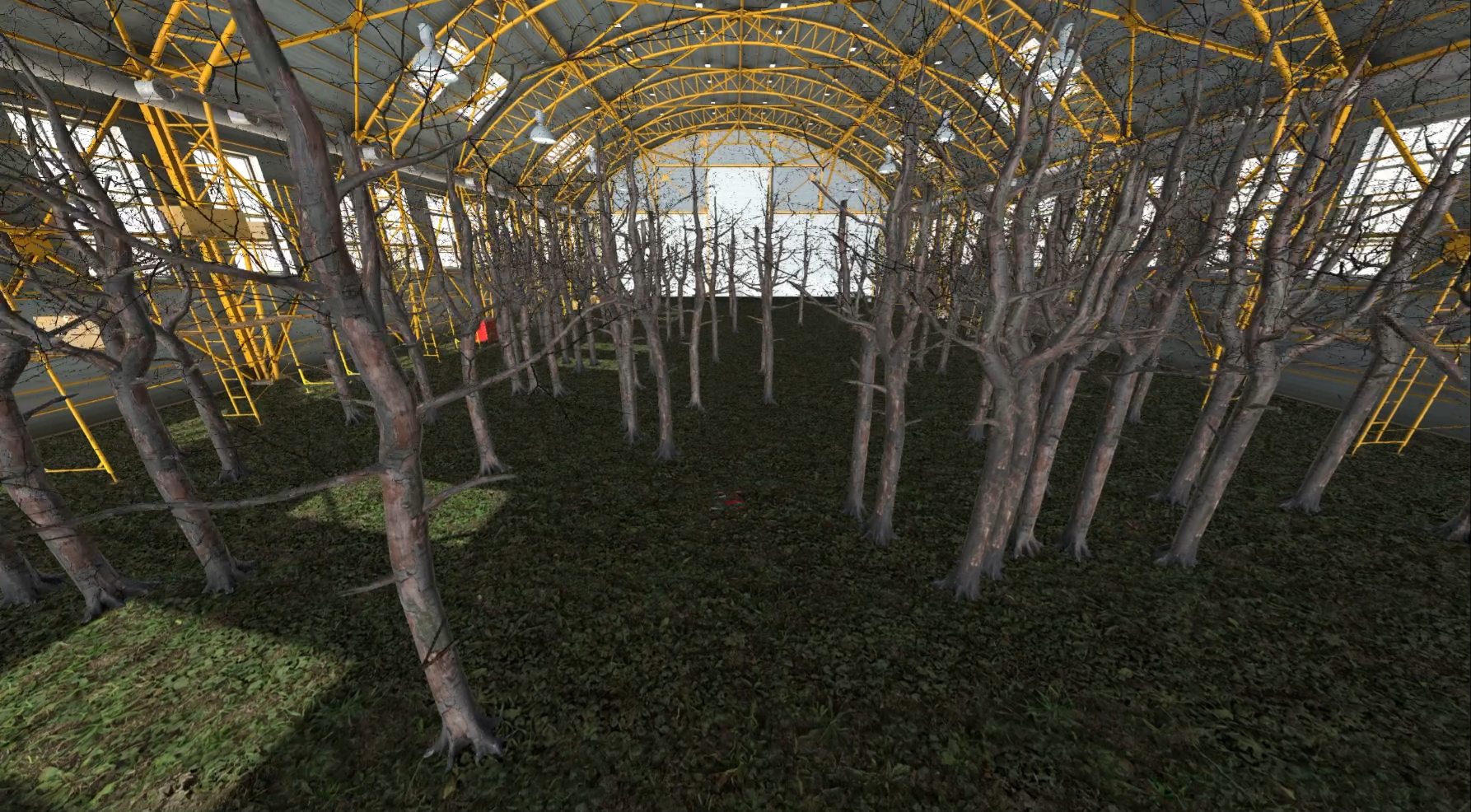} \\
    \footnotesize 
    \qquad (a) Forest   
  }
  \hfill
   \parbox[t]{0.3\linewidth}{
    \centering
    \includegraphics[width=0.8\linewidth]{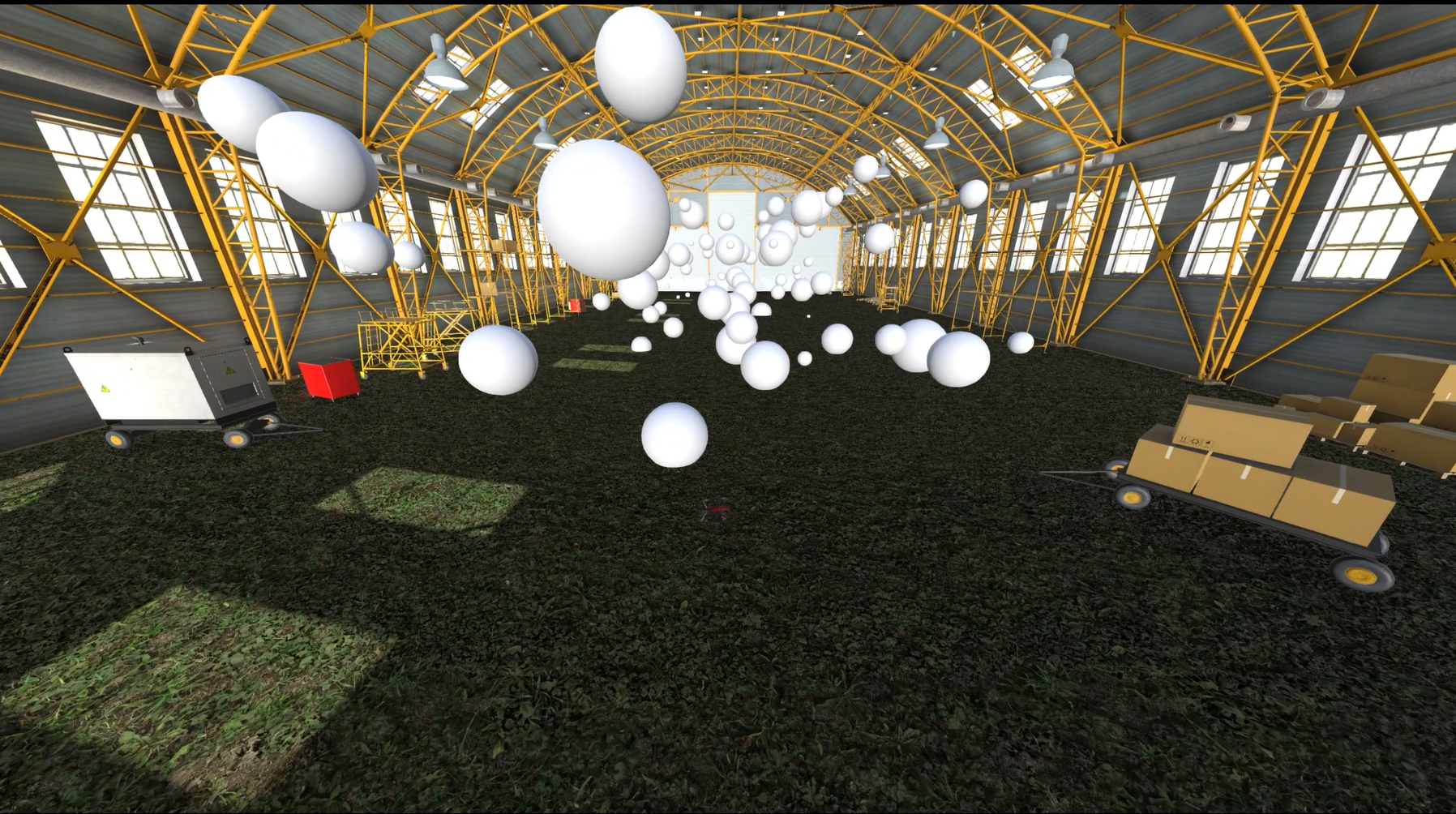} \\
    \footnotesize
    \qquad (b) Static Scene 
  }
  \hfill
  \parbox[t]{0.3\linewidth}{
    \centering
    \includegraphics[width=0.8\linewidth]{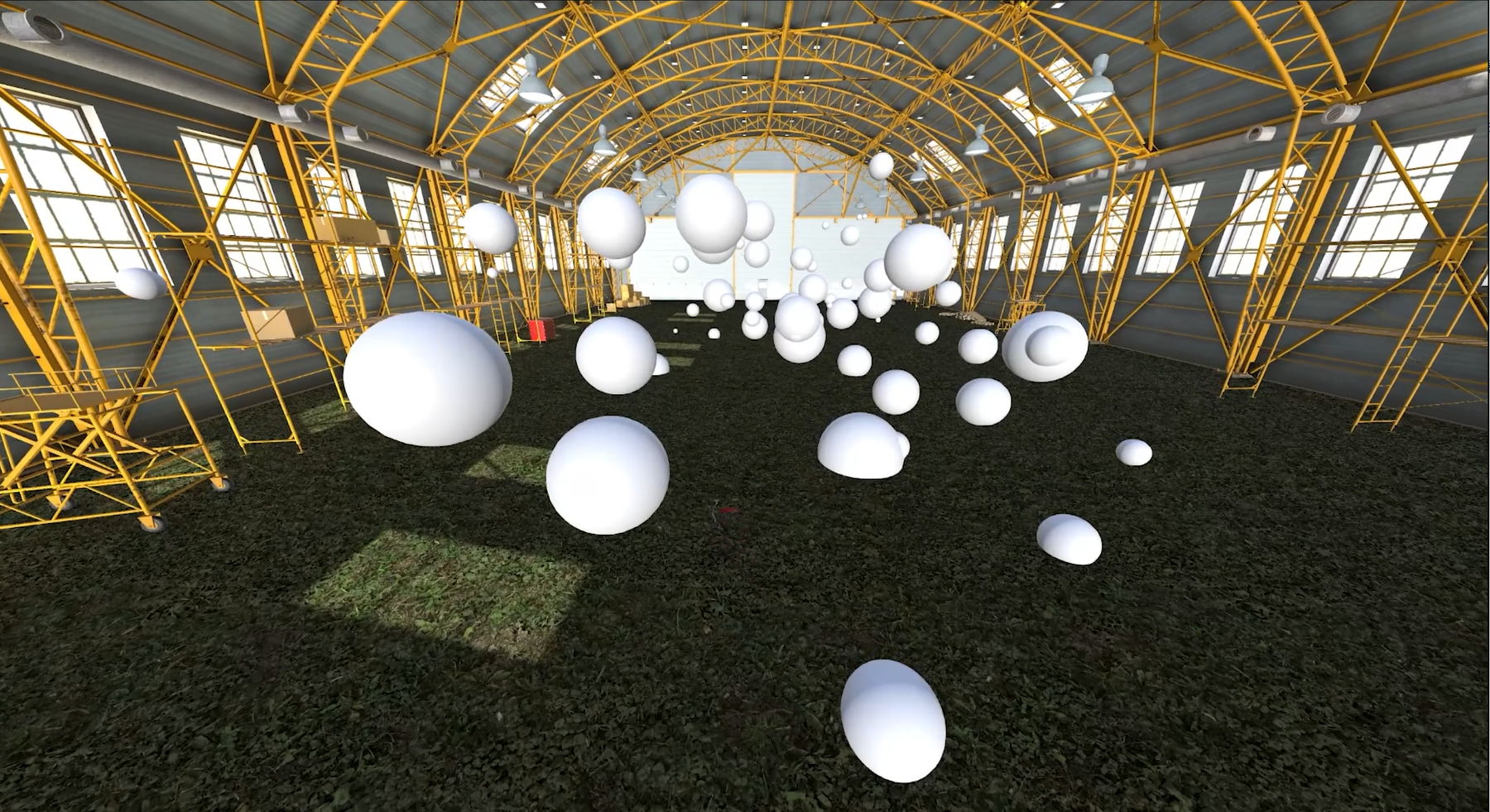} \\
    \footnotesize
    \hspace{2pt} (c) Mixed Scene 
  }
  \caption{Schematic Diagrams of Scenarios with Various Obstacles: (a) Scenarios with Trees; (b) Scenarios with Static Spherical Obstacles; (c) Scenarios with Mixed (Static and Dynamic) Spherical Obstacles;}
  \label{fig:obstacle}
\end{figure*}

\textbf{Bidirectional CrossAttention Feature Fusion}: To effectively fuse depth and event features, we propose a lightweight bidirectional crossattention module (see \Cref{fig:bidirectional_cross-attention_fusion}, i.e., the Feature Fusion part in the \Cref{fig:overall_student_network_architecture}). In contrast to simple concatenation or unidirectional fusion, this module establishes a dual, interactive attention flow between the two modalities:
\begin{figure}[htbp]
      \centering
  \includegraphics[width=0.9\linewidth]{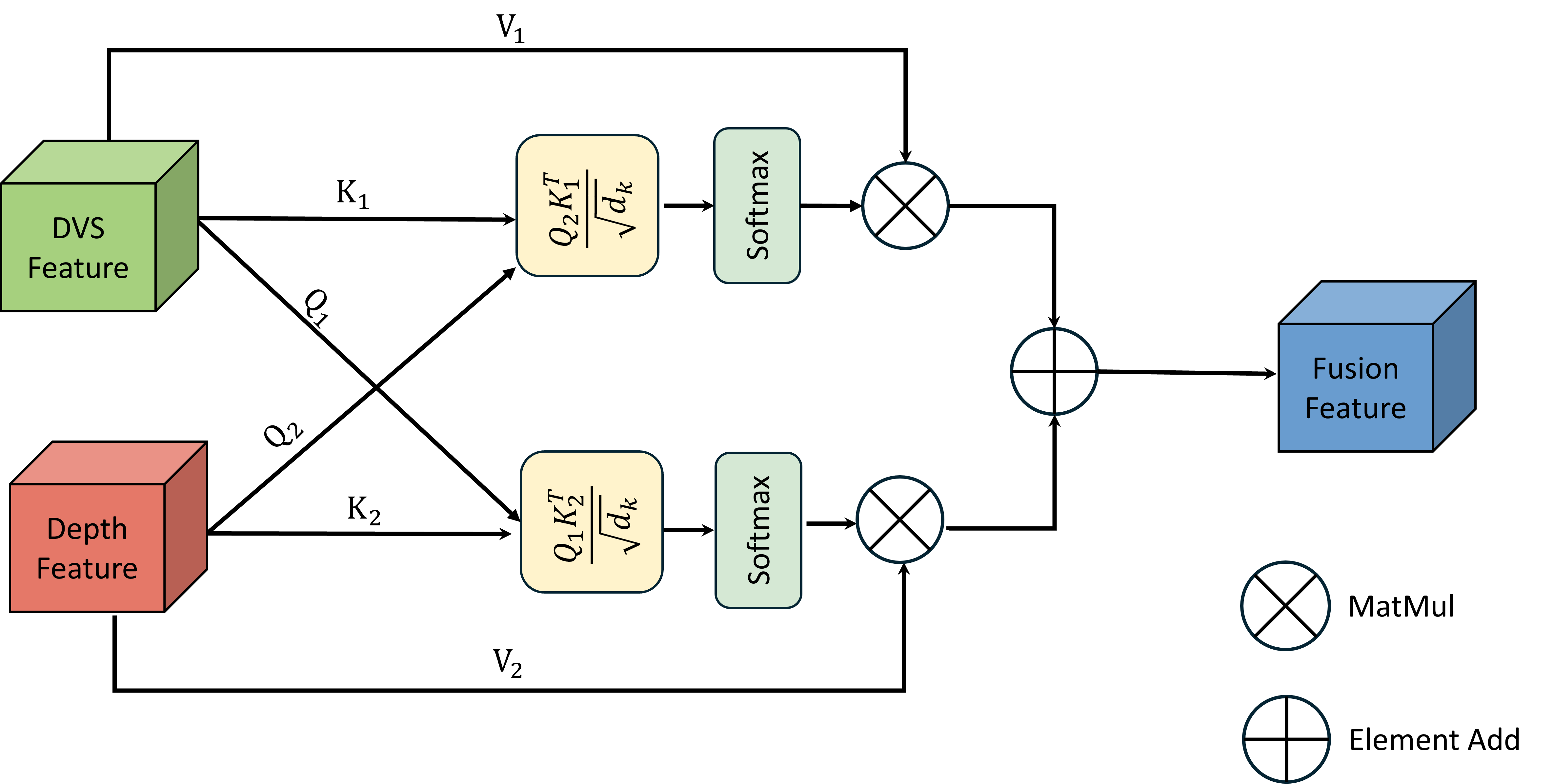}
  \caption{Schematic diagram of the bidirectional attention module}
  \label{fig:bidirectional_cross-attention_fusion}
\end{figure}

\begin{itemize}
    \item \textbf{Depth2event Attention}: Using depth features as the Query $Q_2$ and event features as the Key-Value pair $K_1,V_1$ shown in the \Cref{fig:bidirectional_cross-attention_fusion}, this path anchors spatial static within the dynamic event context. It enables the model to selectively retrieve relevant motion cues, leading to a precise fusion of structural and temporal information.
    \begin{equation}
    \begin{aligned} F_{\mathrm{depth2event}}&=\operatorname{Attention}\left(Q_2, K_1, V_1\right) \\
    &=V_1\operatorname{softmax}\left(\frac{Q_2^T K_1}{\sqrt{d_k}}\right)
    \end{aligned}
    \end{equation}
    \item \textbf{Event2depth Attention}: Using event features as the Query $Q_1$ and depth features as $K_2,V_2$ shown in the \Cref{fig:bidirectional_cross-attention_fusion}, this path enriches dynamic events with the dense spatial structure from the depth map. This completes the bidirectional loop, ensuring motion information is grounded in geometric context.
    \begin{equation}
    \begin{aligned}
    F_{\mathrm{event2depth}}&=\operatorname{Attention}\left(Q_1,K_2, V_2\right)\\
    &=V_2\operatorname{softmax}\left(\frac{Q_1^T K_2}{\sqrt{d_k}}\right)
    \end{aligned}
    \end{equation}
\end{itemize}

The two attention paths are then integrated through element-wise addition,
\Cref{eq:fusion}, providing a balanced fusion of depth and event features with minimal computational overhead

\begin{equation}
\begin{aligned} \label{eq:fusion}
    F_{\text {fusion }}&=F_{\text {depth2event }}+F_{\text {event2depth }}
\end{aligned}
\end{equation}

\textbf{State Encoding \& Command Prediction}: To map the fused visual features to control commands, the UAV’s state is first encoded by an MLP, concatenated with the visual features, and then passed through a second MLP (\Cref{fig:overall_student_network_architecture} MLP part) to output the corresponding velocity command $\tilde{\textbf{v}}_{pred}$.

\textbf{Loss Function}: The loss function consists of two terms: a velocity term (\Cref{eq:loss1}) and a penalty term (\Cref{eq:loss2}). The velocity term is defined as the mean squared error (MSE) between the student's predicted command and the expert's velocity command.

\begin{equation} \label{eq:loss1}
    Loss_{velocity}=||\textbf{v}_{expert} - \tilde{\textbf{v}}_{pred}||_2^2
\end{equation}

The penalty term measures the distance between the predicted trajectory and the nearest obstacle. It is formulated as an exponential penalty function to guide the model toward learning safer obstacle-avoidance behaviors.
\begin{equation} \label{eq:loss2}
    Penalty=\begin{cases}
    2e^{-3d_{near}}, & d_{near} \le 0.5 \\
    0, & d_{near} > 0.5
    \end{cases}
\end{equation}
$d_{near}$ denotes the nearest distance between the path and the obstacle.

The total loss of training can be expressed as \Cref{eq:loss}.
\begin{equation} \label{eq:loss}
    Loss = w_1 Loss_{velocity} + w_2 Penalty
\end{equation}
$w_1=0.7$ and $w_2=0.3$ denote the corresponding weights.

\section{Experiment}

We conduct experiments to answer three questions: (1) Does multimodal fusion outperform single-modality baselines in complex obstacle scenes? (2) Does the \ac{SPS}-based expert improve trajectory quality and success rate compared to grid-based planners? (3) Is bidirectional crossattention superior to unidirectional fusion? The following subsections address these questions respectively.

\subsection{Experimental Setup and Evaluation} 

This section details the experimental setup and evaluation protocol. Our implementation builds upon the ICRA 2022 DodgeDrone Challenge within the Flightmare simulator, focusing on high-speed obstacle avoidance across three scenarios: forest scene (\Cref{fig:obstacle}(a)), static scene (static spherical obstacles, \Cref{fig:obstacle}(b)), and mixed scene (static and dynamic spherical obstacles, \Cref{fig:obstacle}(c))). For training, we used 1,000 trajectories collected from 50 distinct scenes. For testing, we evaluated it in 10 unseen scenes, conducting two flights per scene.

\textbf{Input Preprocessing}: To match the input specifications of the MobileNet-V3-Small model, both the event features and depth features are resized to $224 \times 224$ pixels. The single-channel representations are then expanded along the channel dimension to three channels, yielding an input tensor $I^{224\times 224 \times 3}$. Depth features are further normalized to the range $[0,1]$.

\textbf{Training and Evaluation Setup}: All experiments were performed on a system equipped with an Intel i9-14900KF CPU and an NVIDIA RTX 4090 GPU. The key evaluation metric was the obstacle-avoidance success rate—the percentage of trials in which the \ac{UAVs} successfully flew from start to goal without collision.

\subsection{Overall Obstacle Avoidence Performance}
In this section, we compare against two leading vision-based baselines: event-based (OrigUnet\_lstm~\cite{monocular2024bhattacharyaa}) and depth-based (Vitlstm~\cite{vision2025bhattacharya}).

\begin{figure*}[htbp]
   
      \centering
      \includegraphics[width=0.9\linewidth]{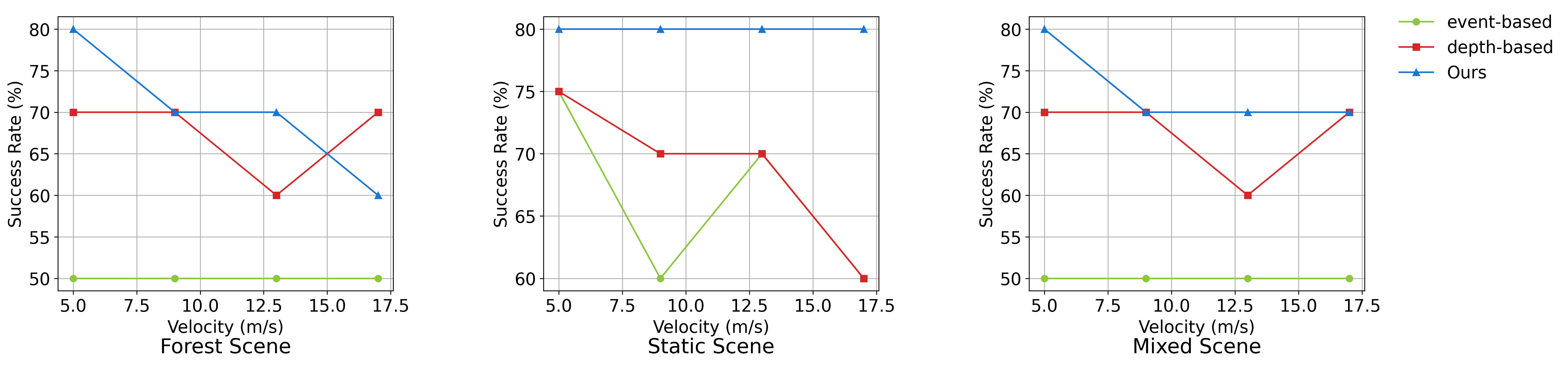}
      \caption{Comparison of fusion models and single-modality baseline models in forest scenes, static spherical obstacle scenes, and mixed spherical obstacle scenes} 
      \label{fig:baseline_comparsion}
\end{figure*}



Our model consistently and substantially outperforms both single-modal baselines across all environments and speeds shown in \Cref{fig:baseline_comparsion}. In forest scenes, our method achieves the highest success rate of 80\% at 5m/s—30\% and 10\% higher than the event-based (50\%) and depth-based (70\%) baselines, respectively. The depth-based model fluctuates and drops to 60\% at 13 m/s, while the event-based model remains flat at 50\%. Our model maintains stable performance between 60\% and 80\% throughout the speed range. In static scenes, our model attains a stable 80\% success rate at all speeds, whereas the two baselines vary between 60\% and 75\%, showing clear degradation and instability. In the most challenging mixed scenes, our model again achieves the top success rate of 80\% at low speed and maintains 70\% at higher speeds (9–17m/s); in contrast, the event-based baseline consistently lags at only 50\%, and the depth-based baseline peaks at 70\% but falls to 60\% at 13m/s. These quantitative comparisons demonstrate that our multimodal fusion strategy not only achieves higher obstacle-avoidance performance but also exhibits significantly greater robustness, effectively overcoming the performance degradation and instability commonly observed in single-modal approaches across static, dynamic, and mixed obstacle environments.

\begin{figure*}[b]
   
      \centering
      \includegraphics[width=0.9\linewidth]{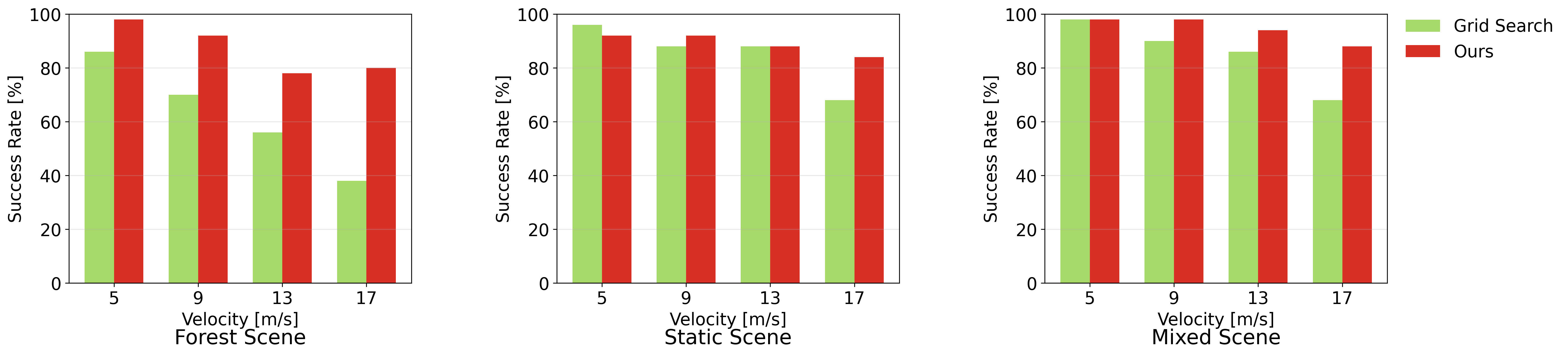}
      \caption{Comparison of obstacle avoidance success rates of expert strategies in different scenes} 
      \label{fig:expert_comparsion}
\end{figure*}


\subsection{Performance of Spherical Principal Search-based Expert Policy} 
In this section, we conduct a comprehensive performance evaluation of the proposed \ac{SPS} against conventional 2D grid-based expert policies. The comparison focuses on three key aspects: computational complexity, avoidance success rate, and trajectory quality analysis.

Compared with the conventional grid-based expert, the \ac{SPS} expert achieves substantial gains in computational efficiency. Traditional experts typically rely on dense sampling over a local 2D grid with quadratic complexity $O(n^2)$ —for instance, evaluating 256 candidate waypoints on a $16 \times 16$ grid—to ensure sufficient coverage of the search space. In contrast, the \ac{SPS} expert constrains candidate waypoints to a spherical manifold defined by the UAV’s current velocity and samples only along principal directions of motion. This achieves linear complexity $O(n)$ by sampling $\frac{n}{2}=8$ points along each of the four principal directions on the sphere. The resulting candidate set contains only 32 waypoints — 12.5\% of the baseline samples, an overall reduction of 87.5\%. The sharp decrease in evaluated waypoints significantly lowers the computational burden on the planning module while still enabling high-quality trajectory selection.

Across all scenes in \Cref{fig:expert_comparsion}, the spherical principal direction search consistently and significantly outperforms the traditional grid search. The advantage is most pronounced at high speeds: at 17m/s, it achieves 80\% success in forest scenes—compared to just 38\% for the baseline—84\% in dynamic scenes (vs. 68\%), and 88\% in static scenes (vs. 68\%). In contrast to the baseline’s sharp degradation, our method exhibits substantially greater robustness, with a much smaller drop in performance as speed increases. These results confirm that our approach not only improves success rates across the board but also maintains high efficiency and reliability under the demanding conditions of high-speed navigation.

\begin{figure}[htbp]
      \centering
      \includegraphics[width=\linewidth]{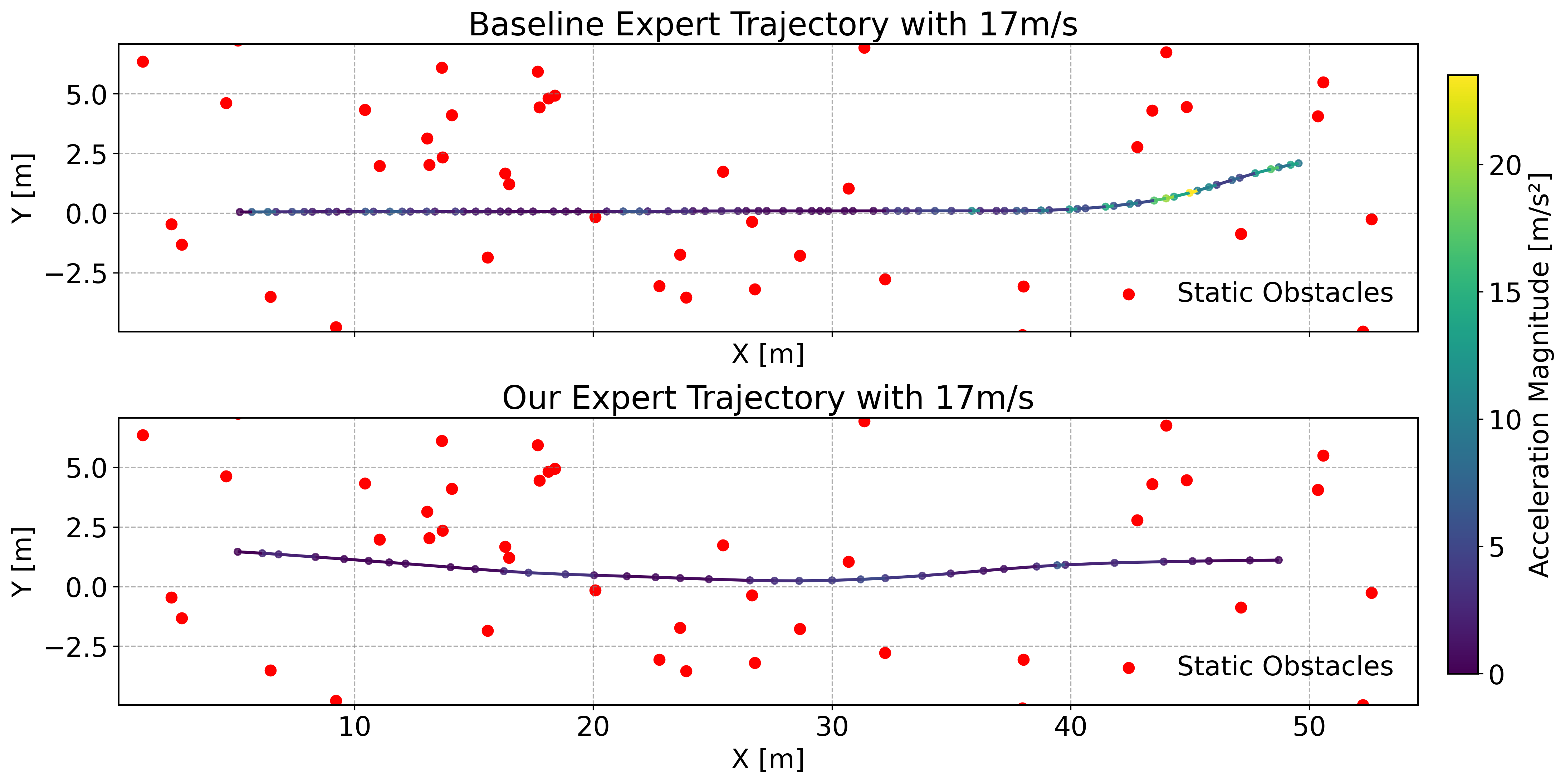}
      \caption{Comparison of expert trajectories in a 17m/s static obstacle scene} 
      \label{fig:expert_traj_comparsion}
\end{figure}

\begin{figure}[htbp]
   
      \centering
      \includegraphics[width=0.8\linewidth]{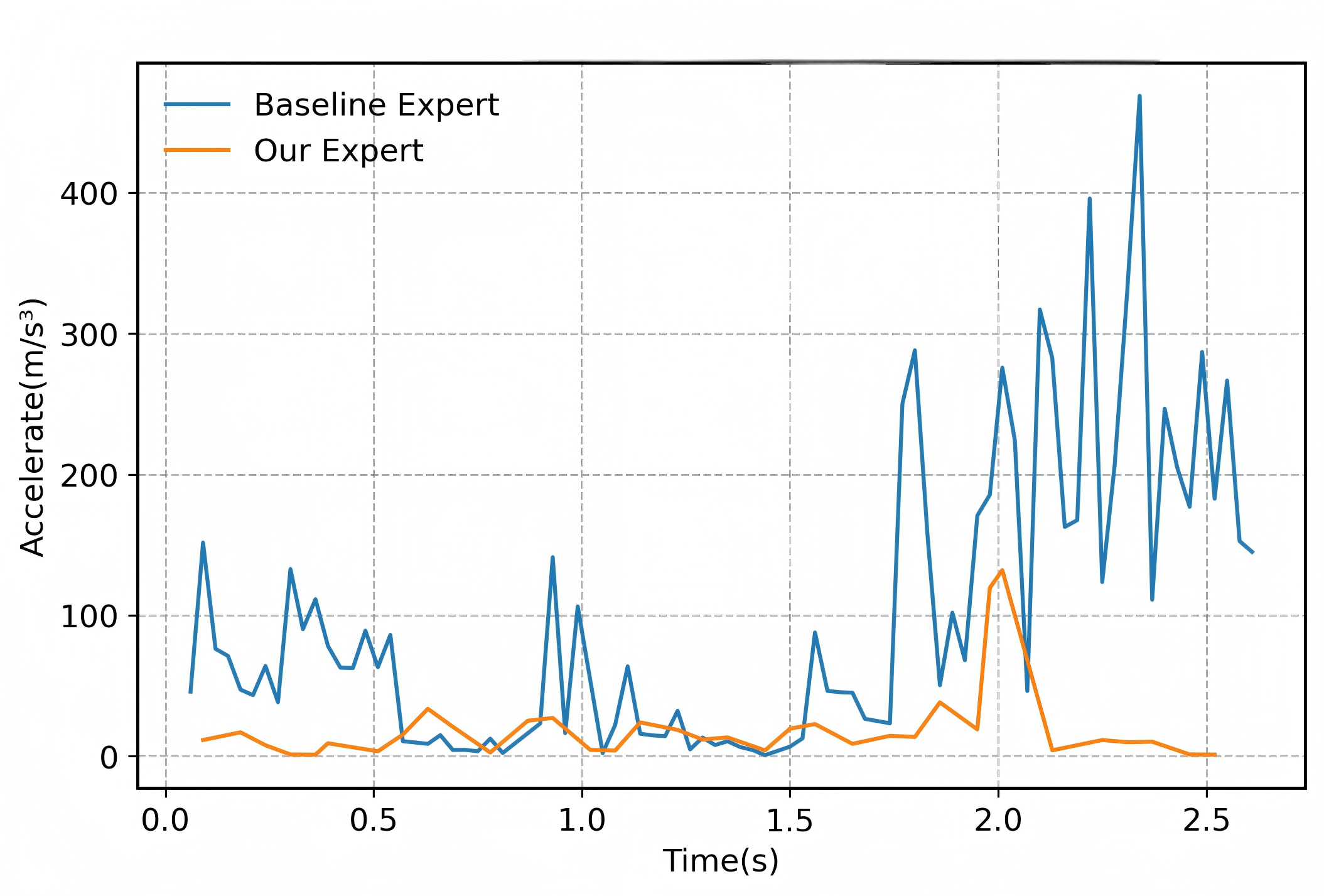}
      \caption{Acceleration Comparison of expert trajectories in a 17m/s static obstacle scene} 
      \label{fig:expert_acceleration_comparsion}
\end{figure}

In addition to higher success rates, our expert demonstrates superior trajectory quality, generating smoother paths with fewer corrections (\Cref{fig:expert_traj_comparsion}). This directly corresponds to significantly lower acceleration. Our expert achieves a maximum of 131.98 m/s² and an average of 19.45 m/s² as shown in \Cref{fig:expert_acceleration_comparsion}, representing reductions of ~72\% and ~80\% from the planar search baseline.

\subsection{Ablation on Bidirectional Crossattention Fusion}
To validate the limitations of unidirectional cross-attention and demonstrate the effectiveness of our bidirectional design, we conducted ablation experiments. To this end, we constructed two baseline models based on unidirectional attention mechanisms: depth2event and event2depth. It is worth noting that typical decision‑level fusion approaches—processing each modality independently and merging results at the task level—are often deployed and evaluated directly on real robotic systems without simulation testing, as seen in prior work \cite{lowlatency2025chen, fastdynamicvision2021hea}.

\textbf{depth2event}: Only depth features can attend to event features (query from depth, key/value from event).

\textbf{event2depth}: Only event features can attend to depth features (query from event, key/value from depth).

\begin{figure}[htbp]
   
      \centering
      \includegraphics[width=0.8\linewidth]{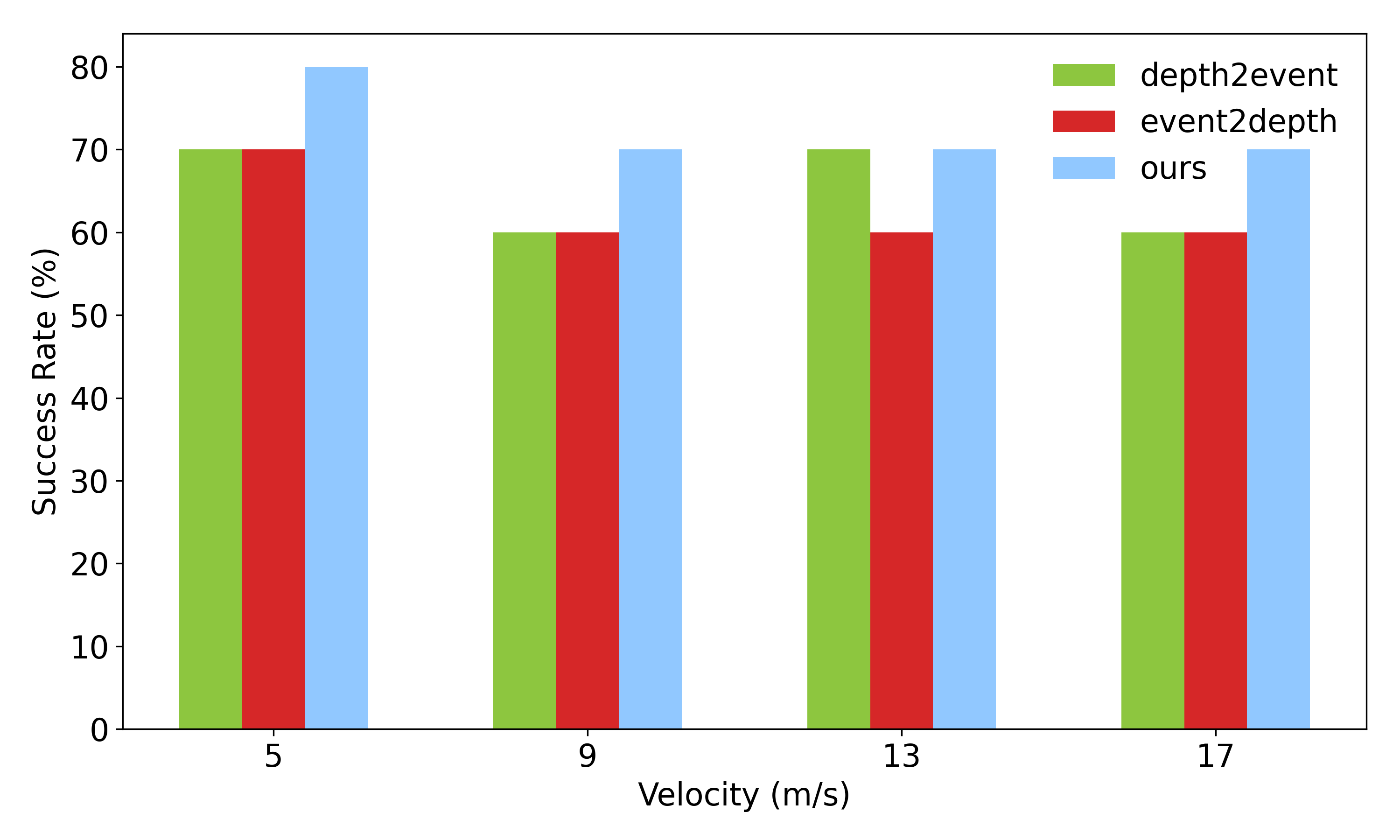}
      \caption{Comparison between bidirectional and single-Direction fusion models in dynamic and static spherical obstacle scenarios} 
      \label{fig:fusion_comparsion}
\end{figure}

The success rates of these model variants in the mixed obstacle scenario are shown in \Cref{fig:fusion_comparsion}. Our model consistently outperforms both unidirectional baselines by margins across all speeds. At 5m/s, ours achieves 80\%, a 10\% improvement over depth2event (70\%) and event2depth (70\%). At 9 m/s, ours maintains 70\%, while both baselines drop to 60\%, yielding a 10\%. At 13 m/s, Ours reaches 70\%, surpassing depth2event (70\%, tied) and event2depth (60\%) by up to 10\%. At the highest speed of 17 m/s, ours remains at 70\%, outperforming both depth2event and event2depth by 10\%. These consistent and substantial gains clearly demonstrate the superiority of bidirectional information exchange.

Our findings further underscore the critical role of this bidirectional design. Both unidirectional variants exhibit pronounced performance degradation as speed increases—depth2event fluctuates between 60\% and 70\%, while event2depth steadily declines to 60\%. This decline highlights the synergistic nature of our fusion approach: static depth information provides essential spatial context for interpreting event data, while high-frequency event streams actively compensate for the motion blur and latency inherent in depth images. Such bidirectional exchange is key to achieving robust multimodal perception during high-speed flight.

\section{CONCLUSIONS}

We propose an end-to-end network that fuses depth and event modalities for high-speed autonomous obstacle avoidance in \ac{UAVs}. To overcome the perceptual limitations of single-sensor systems in mixed scenes, we introduce a lightweight bidirectional crossattention mechanism that integrates depth images with event data to enable complementary perception. We further design an SPS-based expert planner to improve the quality and efficiency of demonstrations for imitation learning, reducing planning complexity from quadratic to linear. Simulation results show that our method generalises robustly across static, dynamic, and mixed obstacle scenes. The bidirectional fusion strategy significantly enhances the reliability of perceptual and decision-making during high-speed flight. Nevertheless, this study has two main limitations: event streams are synthesized from video sequences using Vid2e rather than captured by a physical event camera, and the evaluation is limited to spherical obstacles and forest scenes. Future work includes real-world deployment with actual event cameras and extension to more diverse and densely cluttered environments.






\section*{ACKNOWLEDGMENT}

This work was supported by the National Natural Science Foundation of China (Grant Nos. 62372461 and 62406335).


\printbibliography

\end{document}